\documentclass[10pt,twocolumn]{article} 
\usepackage{simpleConference}
\usepackage{times}
\usepackage{graphicx}
\usepackage{amssymb}
\usepackage{url,hyperref}

\usepackage{tabularx}
\usepackage{soul}
\usepackage{fancybox}
\usepackage{enumitem}

\usepackage{epstopdf}
\usepackage[latin1]{inputenc}
\usepackage{hyperref}
\usepackage{xstring}
\usepackage{url}
\usepackage{here}
\usepackage{amsmath}
\usepackage{authblk}

\newcommand{\Tref}[1]{Table~\ref{#1}}

\newcommand{\fref}[1]{Fig.~\ref{#1}}
\newcommand{\Fref}[1]{Figure~\ref{#1}}
\newcommand{\sref}[1]{Sec.~\ref{#1}}
\newcommand{\Sref}[1]{Section~\ref{#1}}

\def\otoo{\textsc{O-O}}
\def\mtoo{\textsc{M-O}}
\def\otom{\textsc{O-M}}
\def\aotoom{\textsc{Po-OM}}
\def\atoo{\textsc{P-O}}

\def\mat{\textsc{Material}}
\def\matstart{\textsc{Material-Start}}
\def\matintermedium{\textsc{Material-Intermedium}}
\def\matfinal{\textsc{Material-Final}}
\def\matsolvent{\textsc{Material-Solvent}}
\def\matothers{\textsc{Material-Others}}
\def\ope{\textsc{Operation}}
\def\property{\textsc{Property}}
\def\propertytime{\textsc{Property-Time}}
\def\propertytemp{\textsc{Property-Temp}}
\def\propertyrot{\textsc{Property-Rot}}
\def\propertypress{\textsc{Property-Press}}
\def\propertyatmosphere{\textsc{Property-Atmosphere}}
\def\propertyothers{\textsc{Property-Others}}
\def\condition{\textsc{Condition}}
\def\next{\textsc{Next}}
\def\coref{\textsc{Coreference}}

\begin{document}

\title{Annotating and Extracting Synthesis Process\\of All-Solid-State Batteries from Scientific Literature}

\author[1,3]{Fusataka Kuniyoshi}
\author[2,3]{Kohei Makino}
\author[1,3]{Jun Ozawa}
\author[2,3]{Makoto Miwa}
\affil[1]{Panasonic Corporation}
\affil[2]{Toyota Technological Institute}
\affil[3]{National Institute of Advanced Industrial Science and Technology (AIST)}

\maketitle
\thispagestyle{empty}

\begin{abstract}
The synthesis process is essential for achieving computational experiment design in the field of inorganic materials chemistry. In this work, we present a novel corpus of the synthesis process for all-solid-state batteries and an automated machine reading system for extracting the synthesis processes buried in the scientific literature. We define the representation of the synthesis processes using flow graphs, and create a corpus from the experimental sections of 243 papers. 
The automated machine-reading system is developed by a deep learning-based sequence tagger and simple heuristic rule-based relation extractor. Our experimental results demonstrate that the sequence tagger with the optimal setting can detect the entities with a macro-averaged F1 score of 0.826, while the rule-based relation extractor can achieve high performance with a macro-averaged F1 score of 0.887.
\end{abstract}

\section{Introduction}
With the rapid progress in the field of inorganic materials, such as the development of all-solid-state batteries (ASSBs) and solar cells, several materials researchers have noted the importance of reducing the overall discovery and development time by means of computational experiment design, using the knowledge of published scientific literature~\cite{agrawal2016,Butler2018MachineLF,wei2019}. To achieve this, automated machine reading systems that can comprehensively investigate the synthesis process buried in the scientific literature is necessary.

In the field of organic chemistry, a corpus has been proposed in which chemical substances, drug names, and their relations are structurally annotated in documents such as papers, patents, and medical documents, while composition names are provided in the abstracts of molecular biology papers~\cite{Krallinger2015}. Linguistic resources are available in abundance, such as the GENIA corpus~\cite{Kim2003GENIAC} of biomedical events on biomedical texts and the annotated corpus~\cite{kulkarni2018wetlab} of liquid-phase experimental processes on biological papers. In biomedical text mining, the detection of semantic relations is actively researched as a central task~\cite{miwa2012boosting,Scaria2013LearningBP,Berant2014ModelingBP,Rao2017BiomedicalEE,Rahul2017BiomedicalET,Bjrne2018BiomedicalEE}. However, the relations in biomedical text mining represent the cause and effect of a physical phenomenon among two or more biochemical reactions, which differs from the procedure of synthesizing materials.

In the field of inorganic chemistry, only several corpora have been proposed in recent years. A general-purpose corpus of material synthesis has been built for inorganic material by aligning the phrases extracted by a trained sequence-tagging model~\cite{Kononova2019TextminedDO}. However, this corpus did not include relations between operations, and therefore, it was difficult to extract the step-by-step synthesis process.
While, an annotated corpus has been created with relations between operations for synthesis processes of general materials such as solar cell and thermoelectric materials~\cite{mysore-etal-2019-materials}. However, the synthesis processes of ASSBs are hardly included even though the operations, operation sequences, and conditions also have differences due to the characteristics of the synthesis process for each material category.

In this study, we took the first step towards developing a framework for extracting synthesis processes of ASSBs. We designed our annotation scheme to treat a synthesis process as a synthesis flow graph, and performed annotation on the experimental sections of 243 papers on the synthesis process of ASSBs. The reliability of our corpus was evaluated by calculating the inter-annotator agreements. We also propose an automatic synthesis process extraction framework for our corpus by combining a deep learning-based sequence tagger and simple heuristic rule-based relation extractor. A web application of our synthesis process extraction framework is available on our project page~\footnote{\url{http://synth-extractor-demo.fusataka-k.com/}}. We hope that our work will aid in the challenging domain of scholarly text mining in inorganic materials science.

The contributions of our study are summarized as follows:
\begin{itemize}
    \item We designed and built a novel corpus on synthesis processes of ASSBs named SynthASSBs, which annotates a synthesis process as a flow graph and consists of 243 papers.
    \item We propose an automatic synthesis process extraction framework by combining a deep learning-based sequence tagger and rule-based relation extractor. The sequence tagger with the best setting detects the entities with a macro-averaged F1 score of 0.826 and the rule-based relation extractor achieves high performance with a macro-averaged F1 score of 0.887 in macro F-score.
\end{itemize}

\section{Annotated Corpus}
In this section, we present an overview of our annotation schema and annotated corpus, which we named the SynthASSBs corpus.

\subsection{Synthesis Graph Representation}
\label{sec:graph}

We used flow graphs to represent the step-by-step operations with their corresponding materials in the synthesis processes. Using the flow graphs, it was expected that we could represent links that are not explicitly mentioned in text. 
In the inorganic materials field, there is a representation of the synthesis process using a flow graph and the definition of annotation labels in experimental paragraphs~\cite{KIM20198}. In our annotation scheme, we followed their definition, with three improvements: (1) the property of an operation is treated as a single phrase, and not as a combination of numbers and units, (2) each label has been modified to capture the conditions necessary to synthesize the ASSB, and (3) a relation label for coreferent phrases is included to understand the anaphoric relations.
A flow graph for the ASSB synthesis process is represented by a directed acyclic graph $G = (V, E)$, where $V$ is a set of vertices and $E$ is a set of edges. We provide an example section of the paper~\cite{BAI20161045} in \Fref{fig:text}, and the graph extracted from the sentences in the section in \Fref{fig:graph}.

\subsection{Label Set}
\label{sec:label}
The labels contain vertices and edges for the synthesis graph representation in \Sref{sec:vertices}, and \Sref{sec:edges}, respectively.

\subsubsection{Vertices}
\label{sec:vertices}

% 1-s2.0-S0013468616324045-main
\begin{figure}[t]
 \centering
  \fbox{\begin{minipage}{.80\linewidth}
 \footnotesize
The pure \underline{Li$_4$Ti$_5$O$_{12}$} material, denoted \underline{LTO}, was obtained from \underline{Li$_2$CO$_3$} (\underline{99.99 \%}, \underline{Aladdin}) and anatase \underline{TiO$_2$} (\underline{99.8 \%}; \underline{Aladdin}) precursors, \underline{mixed}, respectively, in a \underline{4:5 molar ratio of Li:Ti}. The precursors, \underline{dispersed} in \underline{deionized water}, were \underline{ball-milled} for \underline{4 h} at a grinding speed of \underline{350 rpm}, and then \underline{calcined} at  \underline{800 $^\circ$C} for \underline{12 h} after \underline{drying}.
\end{minipage}}
\caption{Example of a synthesis process. The underlined phrases relate to the material synthesis process.}
\label{fig:text}
\end{figure}

\begin{figure}[t]
 \centering
 \includegraphics[width=.95\linewidth]{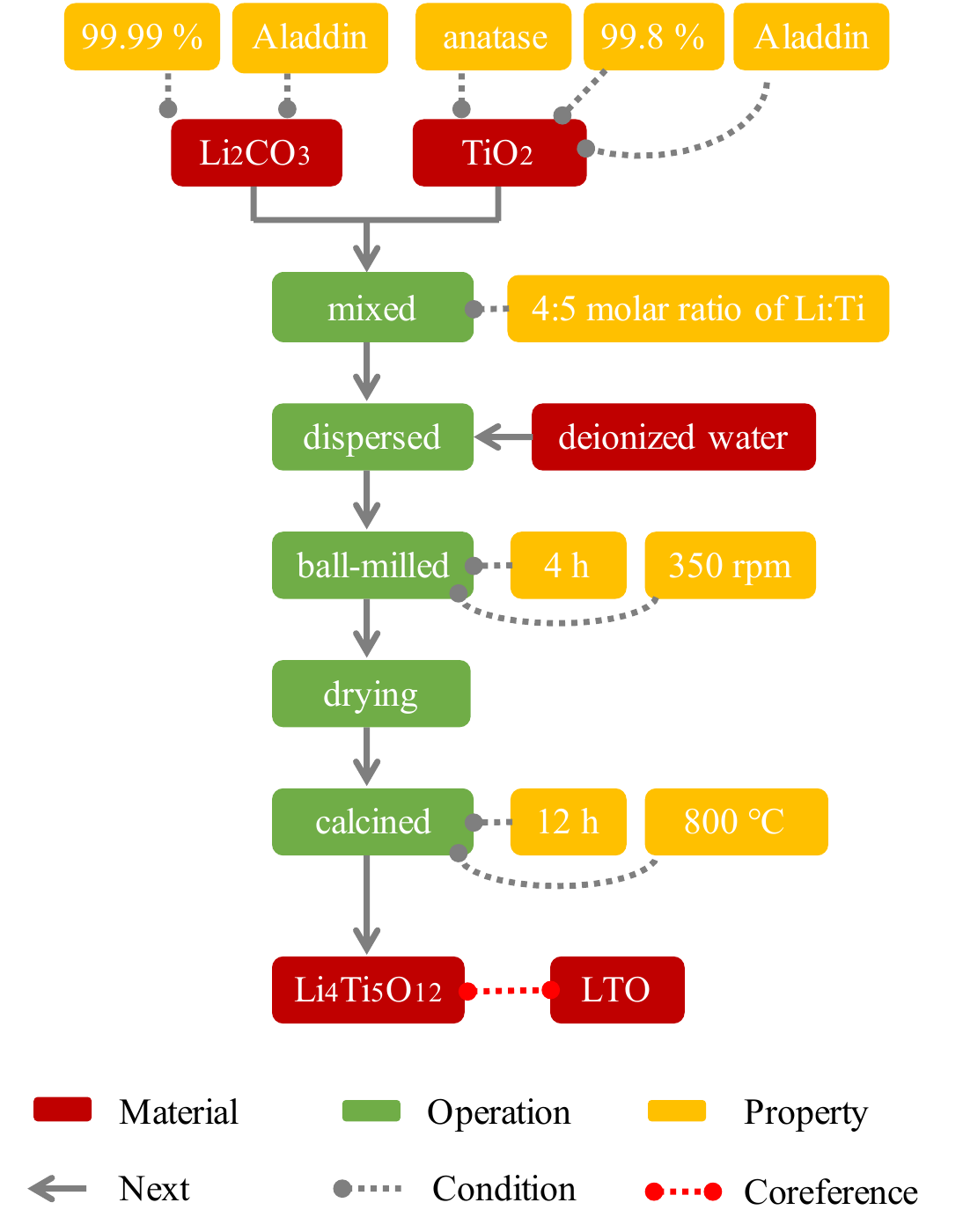}
 \caption{Example of the synthesis graph generated from \Fref{fig:text}. }
 \label{fig:graph}
\end{figure}

\begin{figure*}[t]
 \centering
 \includegraphics[width=\linewidth]{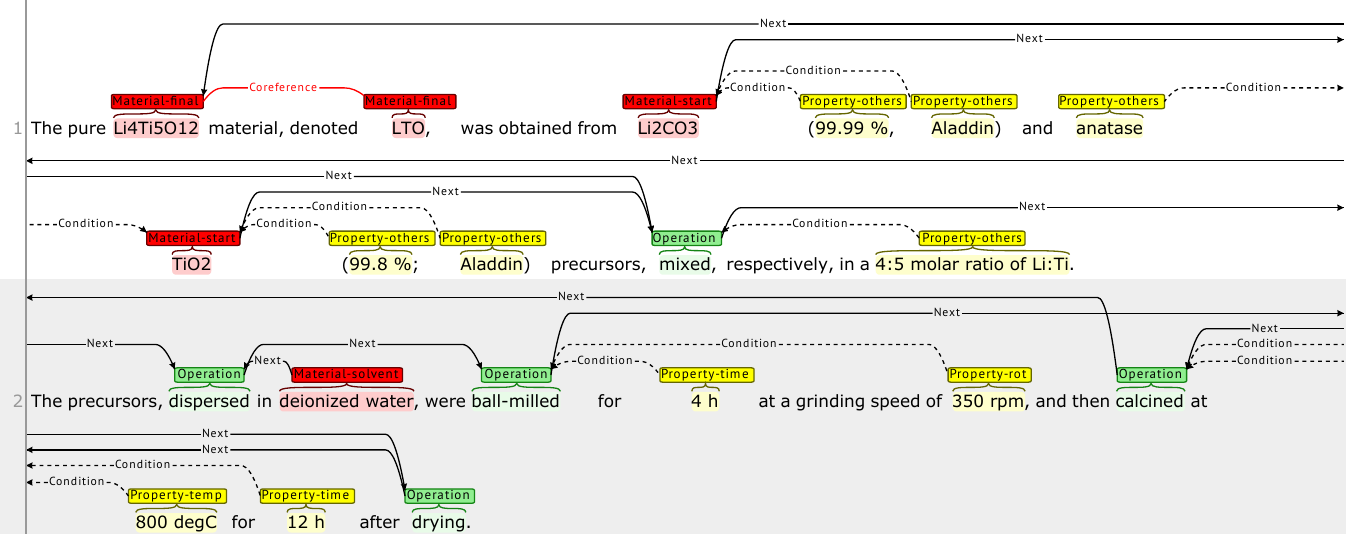}
 \caption{Screenshot of brat interface annotating synthesis process in \Fref{fig:text}. }
 \label{fig:interface}
\end{figure*}

The following vertex labels were defined to annotate spans of text, which correspond to vertices in the synthesis graph. The labels represent the materials, operations, and properties. For the material labels, we labeled all phrases that represent materials in the text, while operation and property labels were added only to those phrases related to the synthesis process. We segmented the roles of materials into categories. Moreover, we introduced multiple property types for analyzing the structure of the synthesis process. The defined labels and their examples mostly taken from \fref{fig:text} are explained in the following.

\textbf{\matstart{}} is a raw material used to synthesize the final material; for example, Li$_2$CO$_3$ or TiO$_2$.

\textbf{\matintermedium{}} indicates an intermediate material produced during the synthesis process; for example, ``then, LixMo$_4$O$_6$ was obtained from a mixture of \underline{InMo$_4$O$_6$} and LiI.''

\textbf{\matfinal{}} represents the final material (or products) of the material synthesis process; for example, Li$_4$Ti$_5$O$_{12}$.

\textbf{\matsolvent{}} is liquid that is used to dissolve substances and create solutions; for example, deionized water, ethanol, or methanol.

\textbf{\matothers{}} represents other materials that are not related to the synthesis process, such as compounds for thin films or catalysts; for example, ``... and then purified with \underline{activated carbon} and \underline{acid alumina}.''

\textbf{\ope{}} represents an individual action performed by the experimenters. It is often represented by verbs; for example, ``... were \underline{ball-milled} for 4 h ...''

\textbf{\propertytime{}} represents a time condition associated with an operation; for example, ``... were ball-milled for \underline{4 h}...''

\textbf{\propertytemp{}} represents a temperature condition associated with an operation; for example, ``... and then calcined at \underline{800 $^\circ$C} ...''

\textbf{\propertyrot{}} indicates a rotational speed condition associated with an operation; for example, ``... at a grinding speed of \underline{350 rpm} ...''

\textbf{\propertypress{}} represents a pressure condition associated with an operation; for example, ``The powder was uniaxially cold pressed at \underline{300 MPa}.''

\textbf{\propertyatmosphere{}} represents an atmosphere condition associated with an operation; for example, ``... was conducted in \underline{Ar atmosphere} for 3 h.''

\textbf{\propertyothers{}} represents other conditions associated with an operation or the manufacturer names and purity associated with a material; for example, ``MgO (\underline{purity 99.999\%}),'' ``... pressed into pellets (\underline{10 mm diameter}, \underline{1 mm thick}),'' and ``the starting materials in the \underline{1/4 molar ratio}.''

\subsubsection{Edges}
\label{sec:edges}
We defined the following three edge labels, which represent the relations between vertices.

\textbf{\condition{}} indicates the conditions of an operation and properties of a raw material (for example, the temperature, time, and atmosphere) for performing an operation. This label is also used to express the relations between a raw material and its manufacturer name or purity.

\textbf{\next{}} represents the order of an operation sequence and indicates the input or output relations between a material and an operation.

\textbf{\coref{}} is a link that associates two or more phrases when these phrases refer to the same material.

\section{Annotation Details and Evaluation}
In this section, we explain the annotation details, including the text preparation, preprocessing, and annotation settings; thereafter we present the settings and results of the inter-annotator agreement experiments.

\subsection{Annotation Details}

We constructed a corpus including the experimental sections of 243 papers on material synthesis processes in the following manner.

We collected papers on experimental processes from online journals.
To limit the annotation target to the ASSB, which is synthesized using the ``solid phase method'' or ``liquid phase method''. We set the search queries to identify papers containing ``solid electrolyte'' or ``ionic conductivity'', but not containing ``poly'', ``SEI'', and ``solid electrolyte interphase'' in the titles, abstracts, and keywords. The four experts in material science are involved in the choice of the paper journal source and selecting keywords.

Thereafter, we manually selected 243 papers that were confirmed to include the synthesis process in the ``Experimental'', ``Preparation'' or ``Method'' sections, because synthesis processes often appear in these sections.
We applied the PDF Parser\footnote{\url{https://github.com/allenai/science-parse}} to extract text from the downloaded PDF papers. We extracted the texts of the above sections, manually corrected several typos, and unified certain orthographical variants in composition formulae and quantitative expressions. For example, a ``$^\circ$C'' was replaced with the token ``degC''.

Finally, we annotated the synthesis graph on the obtained texts. 
Three annotators, who were master's course students in materials science, were involved in the annotation. Annotator A tagged 77 papers, annotator B tagged 68 papers, and annotator C tagged 98 papers. Finally, one professional in materials science verified the annotations of the three student annotators and corrected the annotation errors. We used the brat annotation toolkit~\cite{Stenetorp2012bratAW} for manual annotation. \Fref{fig:interface} illustrates an annotation interface by brat.

\subsection{Inter-Annotator Agreement}
The agreement calculations were based on whether the spans of the labels were precisely matched the three annotators in materials science on the spans by using 30 randomly selected synthesis processes from the SynthASSBs corpus.
We calculated the agreements using Cohen's kappa. For each pair of two annotators selected from the three annotators A, B, and C, the agreement score was calculated by regarding the labels identified by one annotator as gold and the labels by the other annotator as the prediction, and the average of the scores in two directions was determined. 
For the vertices, we calculated two agreement scores: the agreement score of the spans and types (All), and the agreement score of the types on the spans that were annotated by both annotators (Type). For the edges, we also calculated two agreement scores on the vertices that were annotated by both annotators. One score was calculated by comparing the existence of edges and their types (All), while the other score was calculated by comparing the types on the edges that were annotated by both annotators (Type). The inter-annotator agreement results are presented in \Tref{tab:agree}.
\begin{table}[t!]
  \begin{center}
    \begin{tabular}{lcccc} 
       & \multicolumn{2}{c}{Vertices} & \multicolumn{2}{c}{Edges} \\ \hline
      Annotators & All & Type & All & Type \\ \hline
      A--B & 0.637 & 1.000 & 0.705 & 0.990 \\
      B--C & 0.667 & 1.000 & 0.671 & 0.991 \\
      A--C & 0.608 & 1.000 & 0.651 & 0.990 \\ 
    \end{tabular}
    \caption{Inter-annotator agreement results using Cohen's kappa.}
    \label{tab:agree}
  \end{center}
\end{table}
We confirmed that the types (Type) of vertices and edges were almost perfectly matched among the annotators (both kappa coefficients were over 0.99) and the spans and types (All) of them were also substantially matched. This demonstrates that the annotation scheme of the vertices and edges was clear when selecting types. However, the kappa coefficients in the All settings were lower than those in the Type settings. This indicates that annotation ambiguity was caused when deciding which phrase should be involved in the synthesis process. We leave the improvements in the annotation guidelines to reduce this ambiguity problem for future work.

\subsection{Statistics}
Several key statistics of the SynthASSBs corpus, such as the number of documents, sentences, tokens, and entities, are summarized in \Tref{tab:stat-corpus}. The number of vertices or edges per type is indicated in \Tref{tab:stat-type}. In the statistics, we used scispaCy~\cite{Neumann2019ScispaCyFA}~\footnote{\url{https://github.com/allenai/scispacy}} to split sentences, perform tokenization and extract entities.

\begin{table}[t!]
  \begin{center}
    \begin{tabular}{lr}
    Item & Count  \\ \hline
        Documents & 243 \\
        Sentences & 2,877 \\
        Tokens & 46,477 \\
        Entities & 10,995 \\
        Vertex types & 12 \\
        Edge types & 3 \\
        Avg. sentences/document & 12 \\
        Avg. tokens/document & 191 \\
        Avg. entities/document & 45 \\ 
        %Avg. sentence length/sentence & 80 \\
    \end{tabular}
    \caption{SynthASSBs corpus statistics.}
  \label{tab:stat-corpus}
  \end{center}
\end{table}

\begin{table}[t!]
  \begin{center}
    \begin{tabular}{lr}
    Vertex / Edge types & Count  \\ \hline
        \mat{} & 2,749 \\
        \matstart{} & 1,319 \\
        \matintermedium{} & 138 \\
        \matfinal{} & 532 \\
        \matsolvent{} & 212 \\
        \matothers{} & 548 \\
        \ope{} & 1,680 \\
        \property{} & 3,994 \\
        \propertytemp{} & 704 \\
        \propertytime{} & 642 \\
        \propertyrot{} & 66 \\
        \propertypress{} & 81 \\
        \propertyatmosphere{} & 275 \\
        \propertyothers{} & 2,226 \\
        \condition{} & 4,139 \\
        \next{} & 3,018 \\
        \coref{} & 759 \\
        TOTAL & 23,082 \\
    \end{tabular}
    \caption{Statistics of vertices and edges annotated in SynthASSBs corpus.}
  \label{tab:stat-type}
  \end{center}
\end{table}

\section{Synthesis Process Extraction}
\begin{figure}[t!]
 \centering
 \includegraphics[width=\linewidth]{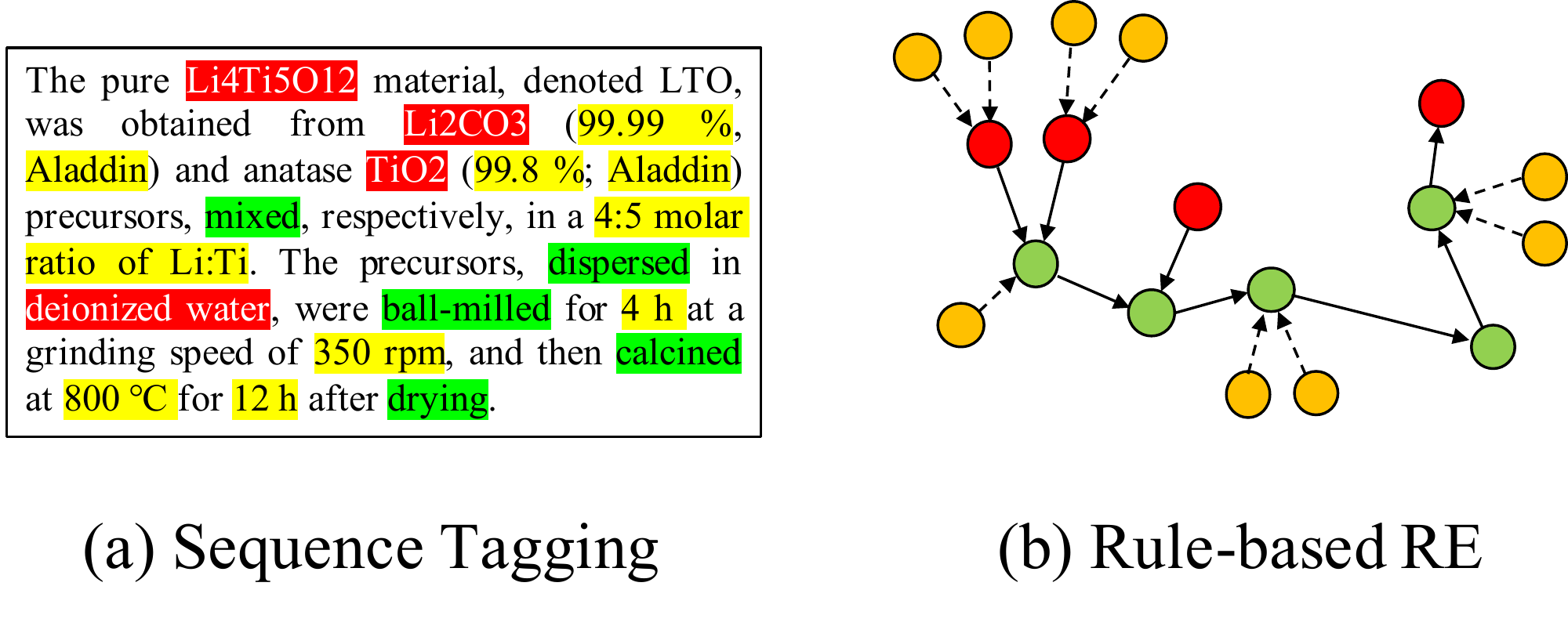}
 \caption{Overview of synthesis process extraction. The red phrases and circles indicate terms related to materials, green indicates operations, and yellow indicates properties. The solid and broken arrows represent the next and condition edges, respectively.}
 \label{fig:framework}
\end{figure}

Our framework performed extraction of synthesis processes in a pipeline manner, using two modules: deep learning-based sequence taggers for extracting the phrases we defined as vertices, and a rule-based relation extractor (RE) for connecting the edges that were pairs of extracted phrases. As illustrated in \Fref{fig:framework}, our framework first performed sequence tagging (a) to extract the phrases related to the material synthesis process. Thereafter, the relations between entities were extracted by the rule-based RE (b).

\subsection{Sequence Tagging}
\label{sec:ner}
To train the sequence-tagging model, we employed Bi-directional Long Short-Term Memory with Conditional Random Fields ~\cite{Huang2015BidirectionalLM} as a sequence-tagging model to identify the spans of the vertices.
We used six different word representations in the neural network-based sequence tagger: character-level embedding (CE)~\cite{Zhang2015CharacterlevelCN}; byte pair encoding (BPE)~\cite{sennrich-etal-2016-neural}; word embeddings for inorganic material science Mat-WE~\cite{Kim2017MachinelearnedAC} and mat2vec~\cite{Tshitoyan2019}; Mat-ELMo~\cite{Kim2017MachinelearnedAC}, which is an embeddings from language models (ELMo)~\cite{Peters2018} model pretrained on materials science texts; and SciBERT~\cite{Beltagy2019SciBERT}, which is a bidirectional encoder representations from transformers (BERT) model~\cite{Devlin2018BERTPO}, pretrained on biomedical and computer science texts. These representations were fine-tuned during training on the sequence-tagging task.

\subsection{Relation Extraction}
\label{sec:re}
We developed the following five rules using the training portion of the SynthASSBs corpus. 
The illustrations following the rule descriptions are used for visualization. The circles used in the figures represent sequential tokens; the red, green, yellow, and white circles corresponds to \mat{}, \ope{}, \property{}, and other words/phrases, respectively. A bounding box around circles represents a sentence. A solid arrow represents an edge of \next{}, while a broken arrow represents an edge of \condition{}.

\textbf{Rule of \ope{} to \ope{} (\otoo):}

An \ope{} phrase is connected to the next \ope{} phrase in the same sentence or in the next sentences.

\begin{figure}[h!]
 \centering
 \includegraphics[width=0.6\linewidth]{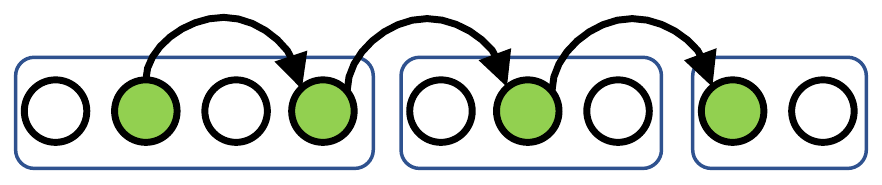}
 \caption{Illustration of \otoo.}
\end{figure}

\textbf{Rule of \mat{} to \ope{} (\mtoo):}

When an \ope{} phrase appears in brackets, a \matstart{} or \matsolvent{} phrase before the left bracket is connected to the \ope{}. In the example sentence ``Samples were prepared from H$_3$BO$_3$, AL$_2$O$_3$, SiO$_2$ and either Li$_2$CO$_3$ (dried at 200 degC),'' the \ope{} phrase ``dried'' written in brackets is connected to its previous \matstart{} phrase, ``Li$_2$CO$_3$'', not ``H$_3$BO$_3$'', ``AL$_2$O$_3$'', and ``SiO$_2$''.

For other \matstart{} or \matsolvent{} phrases, we applied the following rules, ignoring the \ope{} phrases in brackets.
A \matstart{} or \matsolvent{} phrase is connected to its closest \ope{} phrase in a sentence. If two candidates exist within the same distance, the previous candidate is selected.
If no \ope{} phrase exists in a sentence, the phrase is connected to the next-closest \ope{} phrase beyond the sentence boundaries. 
 
\begin{figure}[h!]
 \centering
 \includegraphics[width=0.6\linewidth]{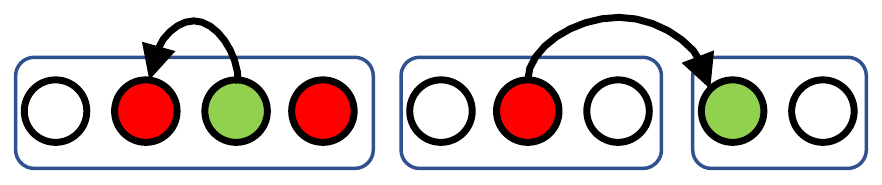}
 \caption{Illustration of \mtoo{}.}
 \label{fig:mst2o}
\end{figure}

\textbf{Rule of \ope{} to \mat{} (\otom):}

An \ope{} phrase that appears at the end of the operation sequence is connected to all \matfinal{} phrases in the text.

\begin{figure}[h!]
 \centering
 \includegraphics[width=0.6\linewidth]{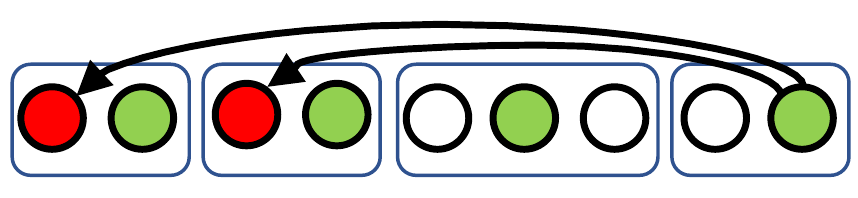}
 \caption{Illustration of \otom.}
\end{figure}

\textbf{Rule of \propertyothers{} to \ope{} or \mat{} (\aotoom):}

When a \propertyothers{} phrase appears in brackets, the phrase is connected to the closet previous \matstart{} phrase. In the example phrase ``TiO$_2$, GeO$_2$ and NH$_4$H$_2$PO$_4$ (purity 99.999 \%),'' ``purity 99.999 \%'' is connected to its closest previous \matstart{} phrase, namely ``NH$_4$H$_2$PO$_4$'', and not ``TiO$_2$'' and ``GeO$_2$''.

A \propertyothers{} phrase is connected to the closest phrase of \matstart{}, \matfinal{}, \matintermedium{}, \matsolvent{}, \matothers{}, or \ope{}. If two candidates exist with the same distance, the previous candidate is selected.

\begin{figure}[h!]
 \centering
 \includegraphics[width=0.6\linewidth]{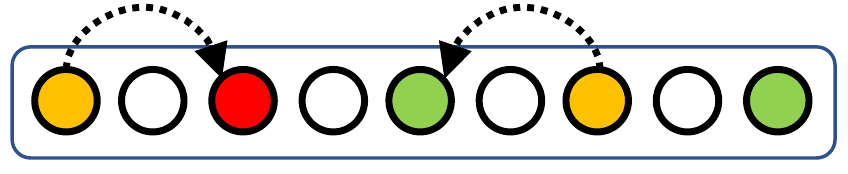}
 \caption{Illustration of \aotoom.}
 \label{fig:ao2om}
\end{figure}

\textbf{Rule of \property{} to \ope{} (\atoo):}

A \propertytime{}, \propertytemp{}, \propertyrot{}, \propertypress{}, or \propertyatmosphere{} (that is, properties other than \propertyothers{}) phrase is connected to its closest previous \ope{} phrase in the sentence or before it. 

\begin{figure}[h!]
 \centering
 \includegraphics[width=0.6\linewidth]{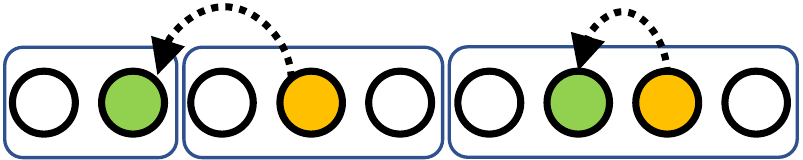}
\caption{Illustration of \atoo.}
 \label{fig:a2o}
\end{figure}

\section{Evaluation}

\begin{table*}[t!]
  \begin{center}
    \begin{tabular}{lcccccccccc} 
       & \multicolumn{3}{c}{\mat{}} & \multicolumn{3}{c}{\ope{}} & \multicolumn{3}{c}{\property{}} & ALL \\ \hline
      Model & F1 & P & R & F1 & P & R & F1 & P & R &  F1\\ \hline
      CE~\cite{Zhang2015CharacterlevelCN} & 0.686 & 0.644 & 0.733 & 0.741 & 0.779 & 0.708 & 0.571 & 0.673 & 0.496 & 0.666\\
      BPE~\cite{sennrich-etal-2016-neural} & 0.860 & 0.837 & 0.883 & 0.799 & \underline{0.785} & 0.814 & 0.706 & 0.726 & 0.686 & 0.788\\
      mat2vec~\cite{Tshitoyan2019} & 0.841 & 0.826 & 0.858 & 0.804 & 0.742 & 0.877 & 0.697 & 0.727 & 0.668 & 0.781\\
      Mat-WE~\cite{Kim2017MachinelearnedAC} & 0.834 & 0.816 & 0.854 & 0.797 & 0.769 & 0.827 & 0.702 & 0.703 & \underline{0.701} & 0.778\\
      Mat-ELMo~\cite{Kim2017MachinelearnedAC} & \textbf{0.917} & \textbf{0.897} & \textbf{0.938} & \underline{0.823} & 0.768 & \textbf{0.887} & \textbf{0.739} & \textbf{0.761} & \textbf{0.718} & \textbf{0.826}\\
      SciBERT~\cite{Beltagy2019SciBERT} & \underline{0.879} & \underline{0.866} & \underline{0.893} & \textbf{0.839} & \textbf{0.798} & \underline{0.884} & \underline{0.709} & \underline{0.749} & 0.673 & \underline{0.809}\\
    \end{tabular}
    \caption{F1 scores of sequence-labeling models with different base representations on development dataset. Macro-averaged F1 scores were calculated using all three coarse-grained types (ALL). The highest and second-highest for each metric are indicated in bold and underline, respectively.}
    \label{tab:result}
  \end{center}
\end{table*}

\subsection{Evaluation Settings}
We evaluated the sequence tagger and rule-based RE individually.
The sequence tagger was implemented using Flair~\cite{akbik-etal-2019-flair}\footnote{\url{https://github.com/zalandoresearch/flair}}, which is a multi-lingual, neural sequence-labeling framework for state-of-the-art natural language processing. When training the sequence tagger, we set the number of training epochs to 200, and used the default hyper-parameters of Flair.

The sequence tagger was evaluated using two settings of type sets.
In the first setting, we extracted three coarse-grained distinct types of vertices in the flow graph: the \mat{}, \ope{}, and \property{} vertices. 
In the second setting, we extracted all 12 fine-grained types of vertices in the flow graph: \matstart{}, \matintermedium{}, \matfinal{}, \matsolvent{}, \matothers{}, \ope{}, \propertytime{}, \propertytemp{}, \propertyrot{}, \propertypress{}, \propertyatmosphere{}, and \propertyothers{}.

We divided the SynthASSBs corpus into three subsets: 145 sections for training, 49 for development, and 49 for testing.
We used an F1 score as the primary evaluation metric. We also report the macro-averaged F1 score of the three coarse-grained types (ALL) for the first setting and the micro-averaged F1 scores for the three coarse-grained types (\mat{}, \ope{}, and \property{}) and the macro-averaged F1 score of these three types (ALL) for the second setting.

We also plot the changes in F1 score of the methods as the training set is increased in increments of 5\% to answer the question about whether the corpus size is large enough to train the sequence tagging. The evaluation was performed on the fine-grained types and the scores were calculated on the development set. We show the micro-averaged F1 scores for the three coarse-grained types and the macro-averaged F1 score of the three types (ALL) for the plot. 

For the rule-based RE, we used 145 sections (used for training in sequence tagging) for designing rules which details showed in \sref{sec:re}, and 98 sections (used as development and testing in sequence tagging) for evaluating the rules.
To evaluate the RE, an F1 score based on an exact match was used as the primary evaluation metric.
We used \coref{} relations in the evaluation: phrase pairs with \coref{} relations were treated as the same phrase in the RE evaluation. 
The performance of the rule-based RE was further analyzed in detail by evaluating the efficiency of the fine-grained labels in the entities as the ablation study, and by demonstrating the accuracy and coverage of each rule.

\subsection{Sequence-Tagging Results}
\label{sec:tag-result}

\Tref{tab:result} summarizes the sequence-labeling results for extracting three coarse-grained vertex types over the six word representations as shown in \sref{sec:ner}. The results show reasonably high performance, in which Mat-ELMo achieved the highest performances, with an F1 score of 0.917 on \mat{}, and 0.826 on ALL, while SciBERT achieved the best score on \ope{}.  

The performance of the sequence tagger with Mat-ELMo, evaluated on the fine-grained types, is presented in \Tref{tab:result-types}. Among the \mat{} types, \matstart{} achieved the highest F1 score of 0.887. The F1 score of \ope{} was 0.821, which was higher than the average. Among the \property{} types, \propertytime{} achieved the highest F1 score of 0.928. However, the F1 score of \matintermedium{} was 0.105 in the sequence tagging, which were 11 phrases incorrectly detected as \matstart{}, and 21 phrases as \matfinal{} out of the 36 phrases. This may be because it is difficult to extract \matintermedium{} without understanding the whole structure of the synthesis process.

\begin{table}[t!]
  \begin{center}
    \begin{tabular}{lccc}
      Types & F1 & P & R\\ \hline
      \mat{} & 0.661 & 0.692 & 0.665\\
      \matstart{} & 0.887 & 0.885 & 0.888\\
      \matintermedium{} & 0.105 & 0.286 & 0.065\\
      \matfinal{} & 0.675 & 0.591 & 0.786\\
      \matsolvent{} & 0.793 & 0.852 & 0.742\\
      \matothers{} & 0.845 & 0.845 & 0.845\\
      \ope{} & 0.821 & 0.792 & 0.852\\
      \property{} & 0.780 & 0.778 & 0.784\\
      \propertytemp{} & 0.842 & 0.806 & 0.880\\
      \propertytime{} & 0.928 & 0.932 & 0.925\\
      \propertyrot{} & 0.889 & 0.857 & 0.923\\
      \propertypress{} & 0.605 & 0.619 & 0.591\\
      \propertyatmosphere{} & 0.775 & 0.775 & 0.775\\
      \propertyothers{} & 0.641 & 0.676 & 0.609\\
      ALL & 0.754 & 0.754 & 0.767\\
    \end{tabular}
    \caption{F1 scores of the sequence labeling models with Mat-ELMo on the development dataset by type. \mat{} and \property{} indicate macro-averaged F1 scores of each fine-grained types, respectively. ALL is the macro-averaged scores of three coarse-grained types (i.e., \mat{}, \ope{}, and \property{} in this table).}
    \label{tab:result-types}
  \end{center}
\end{table}

\begin{figure}[t]
 \centering
 \includegraphics[width=\linewidth]{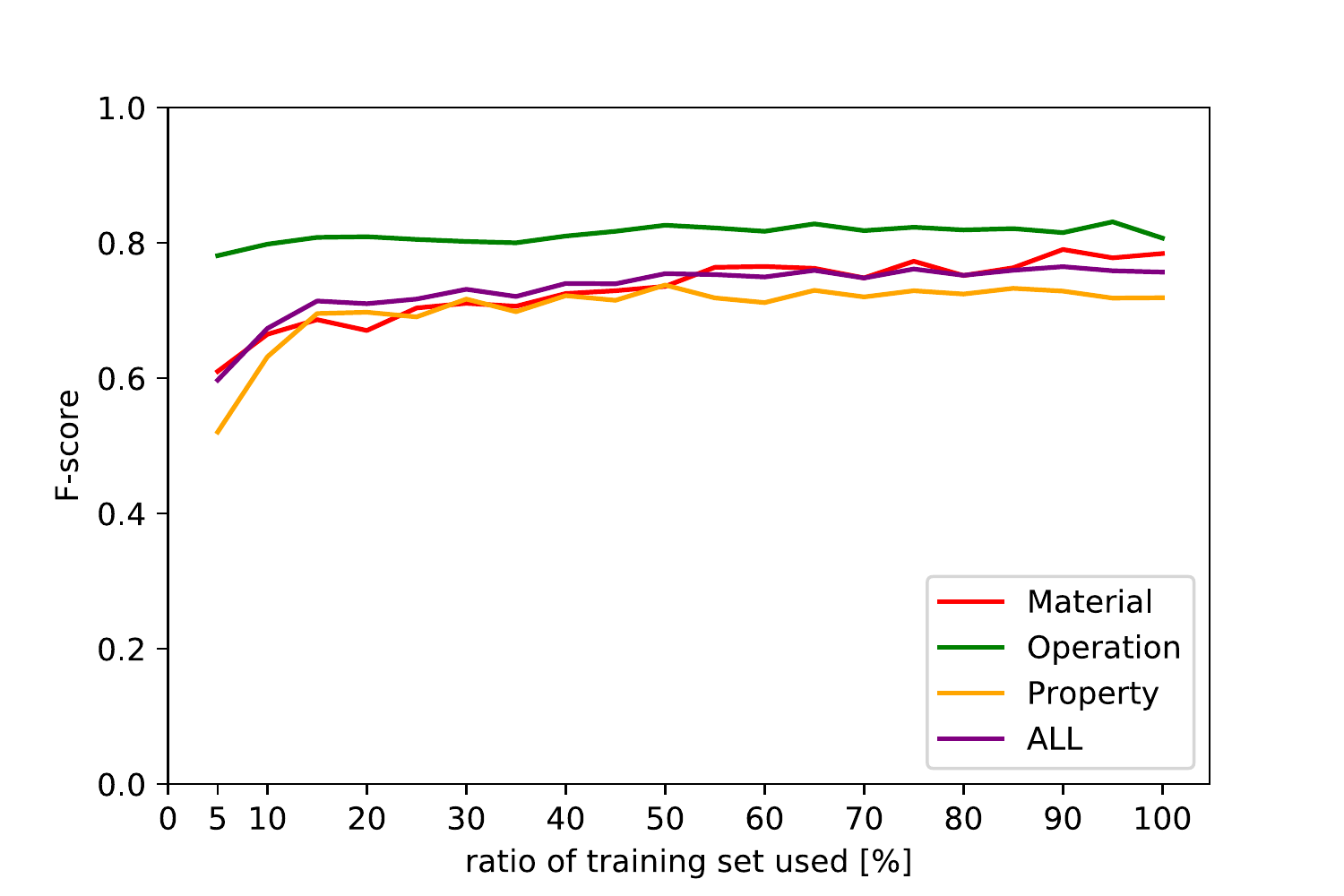}
 \caption{Changes in F1 score according to training set size which is increased in increments of 5 \% until training set size reaches 145 sections. ALL shows the macro-averaged F1 score of the three coarse-grained types.}
 \label{fig:corpus_size}
\end{figure}

Changes in F1 score according to training set size are presented in \Fref{fig:corpus_size}. In this result, we observe that the curves of ALL remain almost flat after using around 20\% of the training set is used. Therefore, we conclude that the size of the SynthASSBs corpus is large enough to train the sequence tagger.
In detail, for \mat{}, the F1 score gradually increases as the training set size increases because material phrases often include unknown terms. \ope{}'s performance is flat after 5\% of the training set is used because there are several types of \ope{} verbs used in the synthesis process. Because \property{} is also steady state when 20\% or more of the training set is used, it seems that the properties are described in a regular manner.

\subsection{Relation Extraction Results}
\Tref{tab:result_ablation} displays the results of the rule-based model as well as the rule-based RE results obtained by the ablation tests.
The high performance with a macro-averaged F1 score of 0.887 shows the effectiveness of the rules.
%\Tref{tab:result_relation} displays the results of the rule-based model and \Tref{tab:result_ablation} presents results of the ablation tests.
To confirm the effectiveness of the fine-grained types or sub-labels, we compared the F1 score with three settings. In the first setting, we extracted the relations without using material sub-labels (-- \mat{}-$\ast$), by applying the rule of \mtoo{} to all of the \mat{} types and ignoring the rule of \otom{}. In the second setting, we extracted relations without using \property{} sub-labels (-- \property{}-$\ast$), by applying the rule of \aotoom{} to all of the \property{} types and ignoring the rule of \atoo{}. The final setting was without either \mat{} or \property{} sub-labels (-- both sub-levels). 
According to the ablation tests, the F1 scores were improved by 7.8\% on \condition{} and 11.1\% on \next{} when applying the sub-label rules.

To analyze the effects of the rules in further detail, the coverage and accuracy for each rule were determined, and these are presented in \Tref{tab:coverage}.
By comparing the rule coverage and accuracy, it could be observed that the rules of \aotoom{} and \atoo{}, which exhibited wide coverage and high accuracy (over 25\% and 85\%, respectively), contributed significantly to the extraction performance. This indicates that the rules for extracting the relation between the \property{} and \mat{} or \ope{} successfully mimicked the manner of reading a paper. 
Although the coverage of the rule \otom{} was extremely low and the accuracy was relatively low (4.6\% and 48.9\%, respectively), this rule was essential for constructing the synthesis graph and could not be omitted.

% \begin{table*}[t]
%   \begin{center}
%     \begin{tabular}{lcccccccc} 
%       & \multicolumn{3}{c}{\condition{}} & \multicolumn{3}{c}{\next{}} &  ALL \\ \hline
%       & F1 & P & R & F1 & P & R & F1 \\ \hline
%       Prediction with gold entities & 0.914 & 0.922 & 0.907 & 0.860 & 0.851 & 0.871 & 0.887 \\
%       Pipeline prediction & 0.253 & 0.289 & 0.226 & 0.274 & 0.249 & 0.305 & 0.264 \\ 
%     \end{tabular}
%     \caption{Rule-based relation extraction result}
%     \label{tab:result_relation}
%   \end{center}
% \end{table*}

\begin{table}[t]
    \centering
    \begin{tabular}{lccc} 
         & \condition{} & \next{} & ALL \\ \hline
         Rule-based RE & 0.914 & 0.860 & 0.887 \\
         -- \mat{}-$\ast$ & 0.914 & 0.749 & 0.832 \\
         -- \property{}-$\ast$ & 0.836 & 0.860 & 0.848 \\
         -- both sub-levels & 0.836 & 0.749 & 0.793 \\
    \end{tabular}
    \caption{F1 scores of rule-based system and ablation test results. Macro-averaged F1 scores were calculated using \condition{} and \next{} (ALL).}
    \label{tab:result_ablation}
\end{table}

% Orientation
% [[ 372  847]
%  [1120    0]]
% R: 0.30517 P: 0.24933 F1: 0.27444
% Relation
% [[ 440 1509]
%  [1083    0]]
% R: 0.22576 P: 0.28890 F1: 0.25346

\begin{table}[t]
    \centering
    \begin{tabular}{lcc} 
     Rule & Coverage & Accuracy \\ \hline
         \otoo{} & 0.219 & 0.811 \\
         \mtoo{} & 0.160 & 0.811 \\
         \otom{} & 0.046 & 0.489 \\
         \aotoom{} & 0.322 & 0.853 \\
         \atoo{} & 0.254 & 0.951 \\ 
    \end{tabular}
    \caption{Coverage and accuracy of our rules applied to training data.}
    \label{tab:coverage}
\end{table}

\section{Qualitative Evaluation}
\begin{figure*}[t]
 \centering
 \includegraphics[width=\linewidth]{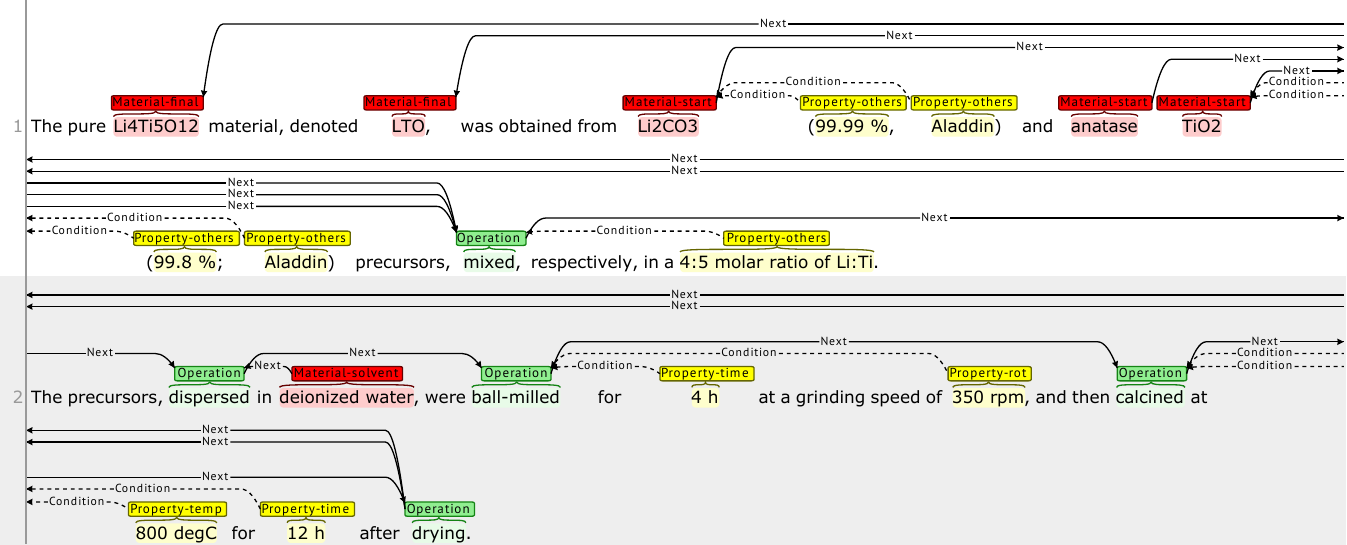}
 \caption{Synthesis process extraction results from the text in \Fref{fig:text}}
 \label{fig:pipeline_prediction}
\end{figure*}

\begin{figure}[t]
 \centering
 \includegraphics[width=.95\linewidth]{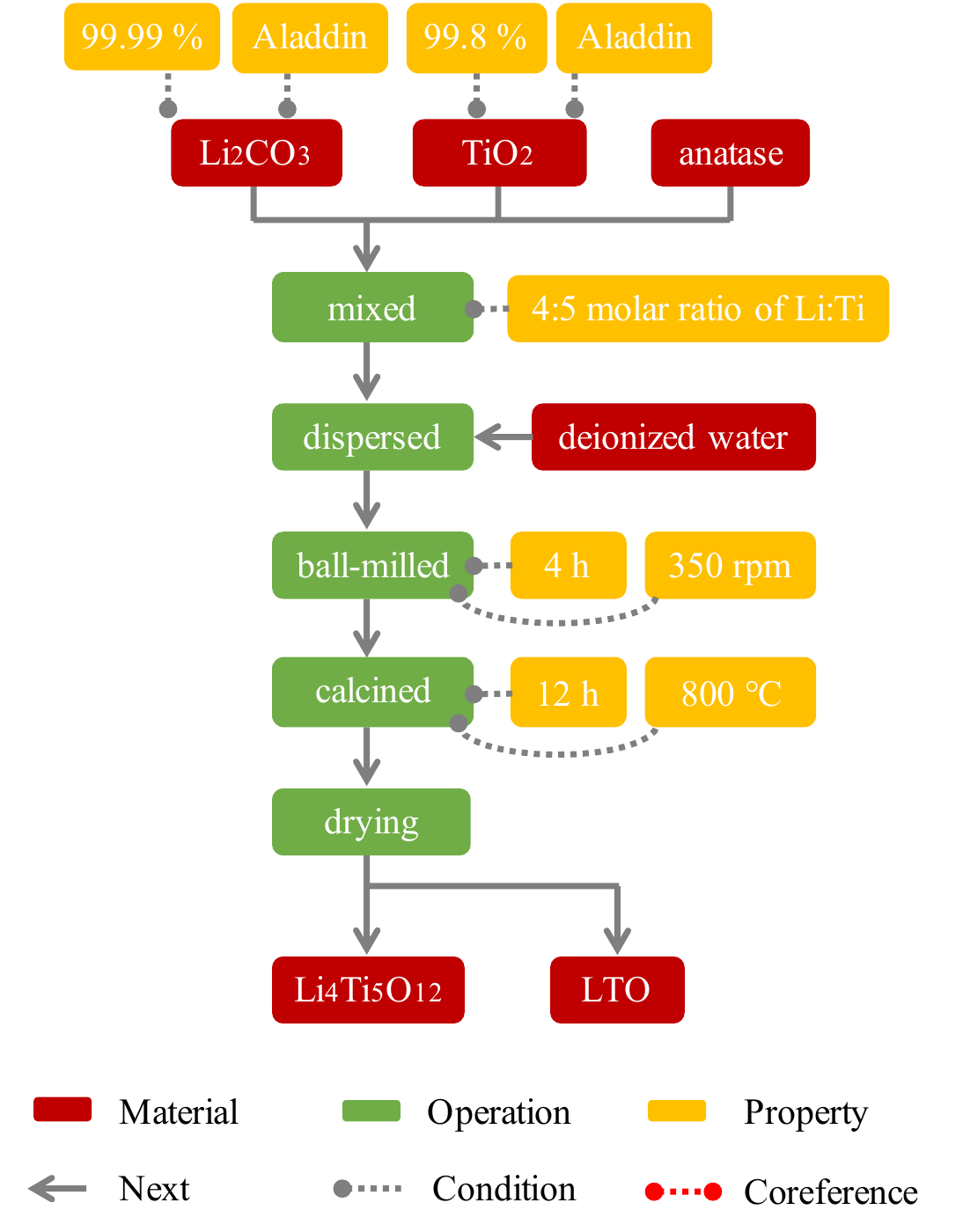}
 \caption{Synthesis graph of the extracted synthesis process in \Fref{fig:pipeline_prediction}. }
 \label{fig:graph_prediction}
\end{figure}

We present a thorough evaluation on a real-world scientific literature to demonstrate the efficacy of our framework. A prediction obtained by our framework and the synthesis graph are shown in \Fref{fig:pipeline_prediction} and \Fref{fig:graph_prediction}, respectively. In this result, our framework could extract phrases related to material synthesis almost without error. In detail, the relations across the sentences were extracted without problems; for example, our framework created a \next{} edge between ``mixed'' written in the first sentence and ``dispersed'' written in the second sentence. Moreover, our framework succeeded in identifying the type of \mat{} even if material written in an abbreviation form; for example, our framework could detect that ``Li$_4$Ti$_5$O$_12$'' and ``LTO'' are \matfinal{} in the first sentence. However, the label type was wrong in ``anatase'' in the first sentence, and \ope{} connection between ``calcined'' and ``drying'' on the second line was different from that labeled by the annotator. This is because our rule-based RE could not understand the meaning of ``after drying''.

\section{Error Analysis}
We analyzed 135 errors in the sequence-tagging results. 
The over-detection errors constituted 49 cases, which were often \propertyothers{} types that were not directly related to the synthesis process; for example, vessel size or thickness, and milling machine properties.
A total of 49 entities were missing and were often caused by \property{} types, were missing due to rare adverbs, adjectives, or units; for example, ``naturally'', ``constant'', ``mm-thick'', and, ``micrometers''.

In the RE, we identified two major problems when we analyzed the 129 errors. The first problem was caused by the definition of the distance, which used the number of words and ignored syntactic structures.
For example, in the sentence ``LiNO3 were weighed according to the stoichiometry of the Li3xLa2/3-xTiO3 and dissolved in ethylene,'' our distance-based rule predicted that ``Li3xLa2/3-xTiO3'' qualifies ``dissolved'' instead of ``weighed''.
This type of problem included 73 errors.
The second problem was complex operation sequences. Where two or more material synthesis processes were described in one document, there were cases in which a synthesis process indicated at the beginning was omitted in the second and subsequent explanations. In such cases, branching and merging of synthesis processes occurred. Our rules assumed that the operation sequence was described sequentially, so they could not identify these processes. This type of complex operation sequence caused 28 errors.
One means of addressing the above problems is to incorporate additional rules; however, it is not realistic to create more rules manually, because the descriptions are sometimes ambiguous, without an understanding of the contents. We are considering developing a deep learning-based extractor that can take syntactic structures into account. 

\section{Related Work}
Process extraction from procedure texts has been studied in a wide range of fields. Such studies include an effort to extract liquid mixing procedures from text~\cite{Long2016SimplerCL}, an annotated corpus of photosynthesis and formation erosion processes~\cite{Dalvi2018TrackingSC}, the extraction of response processes from guidance texts at the time of disaster occurrence~\cite{Guo2018AutomaticEO}, and several attempts to structure and extract a series of cooking-related actions, such as baking and boiling, from cooking recipe sentences~\cite{Mori2014FlowGC,Kiddon2015MiseEP,Maeta2015AFF,Abend2015LexicalEO}.
Numerous language resources exist in the organic chemistry field~\cite{Kim2003GENIAC,Krallinger2015,Tsubaki2017,kulkarni2018wetlab,tanaka-etal-2018-chemical}, which have been annotated with the experimental processes that appear in the papers. Moreover, an attempt has been made to extract processes by applying event extraction methods to realize machine-based text reading for biomedical papers~\cite{miwa2012boosting,Scaria2013LearningBP,Berant2014ModelingBP,Rao2017BiomedicalEE,Rahul2017BiomedicalET,Bjrne2018BiomedicalEE}.
In the inorganic chemistry field, several corpus are available for general-purpose materials~\cite{mysore-etal-2019-materials,Kononova2019TextminedDO}, while some studies are underway to extract the synthesis process from papers~\cite{Mysore2017AutomaticallyEA,Tamari2019PlayingBT}; however, no corpus and extraction system exist for synthesizing ASSBs. Therefore, we have presented a domain specific corpus of the synthesis process for ASSBs, and an automated machine-reading system for extracting the synthesis processes buried in the scientific literature.

\section{Conclusion}\label{sec:conclusion}

This study has addressed the problem of the lack of labeled data, which is a major bottleneck in developing ASSBs. We constructed the novel SynthASSBs corpus, consisting of the experimental sections of 243 papers. The corpus annotates synthesis graphs that represent the synthesis process of ASSBs in text. Moreover, we proposed an automatic synthesis process extraction framework using our corpus by combining a deep learning-based sequence tagger and rule-based relation extractor that mimics the experience in human reading. As a result, the sequence tagger with the best setting can detect the entities with a macro-averaged F1 score of 0.826. Furthermore, the rule-based RE demonstrates high performance with a macro-averaged F1 score of 0.887. 

In future work, we will develop a deep learning-based relation extractor that incorporates syntactic information into the model to improve the extraction performance. We will also apply our extracting framework to existing papers, and, using the extracted abundant knowledge, we will build a computational synthesis design framework for discovering novel material.

\bibliographystyle{abbrv}
\bibliography{main}
\end{document}